\documentclass[conference]{IEEEtran}
\IEEEoverridecommandlockouts
\usepackage{cite}
\usepackage{amsmath,amssymb,amsfonts}
\usepackage{algpseudocode}
\usepackage{graphicx}
\usepackage{textcomp}
\usepackage{xcolor}
\def\BibTeX{{\rm B\kern-.05em{\sc i\kern-.025em b}\kern-.08em
    T\kern-.1667em\lower.7ex\hbox{E}\kern-.125emX}}
\usepackage{amssymb}
\usepackage{amsmath}
\usepackage[ruled,linesnumbered]{algorithm2e}
\usepackage{multirow}
\usepackage{amsmath}
\usepackage{graphics}
\usepackage{makecell}
\usepackage{epsfig}
\usepackage{etoolbox}
\makeatletter
\patchcmd{\@makecaption}
  {\scshape}
  {}
  {}
  {}
\makeatother
\begin{document}

\title{Few-Shot Image Classification via Contrastive Self-Supervised Learning\\
}

\author{\IEEEauthorblockN{Jianyi Li}
\IEEEauthorblockA{\textit{School of Information and Communications} \\
\textit{Xi’an Jiaotong University}\\
Xi'an, P.R. China \\
lijianyi1488@stu.xjtu.edu.cn}
\and
\IEEEauthorblockN{Guizhong Liu}
\IEEEauthorblockA{\textit{School of Information and Communications} \\
\textit{Xi’an Jiaotong University}\\
Xi'an, P.R. China \\
liugz@xjtu.edu.cn }
%\and
% \and
% \and
% \and
 }

\maketitle

\begin{abstract}
Most previous few-shot learning algorithms are based on meta-training with fake few-shot tasks as training samples, where large labeled base classes are required. The trained model is also limited by the type of tasks. In this paper we propose a new paradigm of unsupervised few-shot learning to repair the deficiencies. We solve the few-shot tasks in two phases: meta-training a transferable feature extractor via contrastive self-supervised learning and training a classifier using graph-aggregation, self-distillation and manifold augmentation. Once meta-trained, the model can be used in any type of tasks with a task-dependent classifier training. Our method achieves state-of-the-art performance in a variety of established few-shot tasks on the standard few-shot visual classiﬁcation dataset, with an 8-28\% increase compared to the available unsupervised few-shot learning methods.
\end{abstract}


\section{Introduction}
In recent years deep learning has made major advances in computer vision areas such as image recognition, video object detection and tracking. A deep neural network needs a large amount of labeled data to fit its parameters whereas it is laborious to label so many examples by human annotators. Thus the problem of learning with few labeled samples called few-shot learning has been paid more and more attention. Few-shot learning is described as a classification task set in N -way and k -shot, which means to distinguish N categories, each of which has k (quite small) labeled samples. The model predict classes for new examples only depending on k labeled data. The annotated data is called the support set, and the new data belonging to the N categories is called query set.

\begin{figure}[htbp]
\centerline{\includegraphics[width=10cm, height=4.55cm]{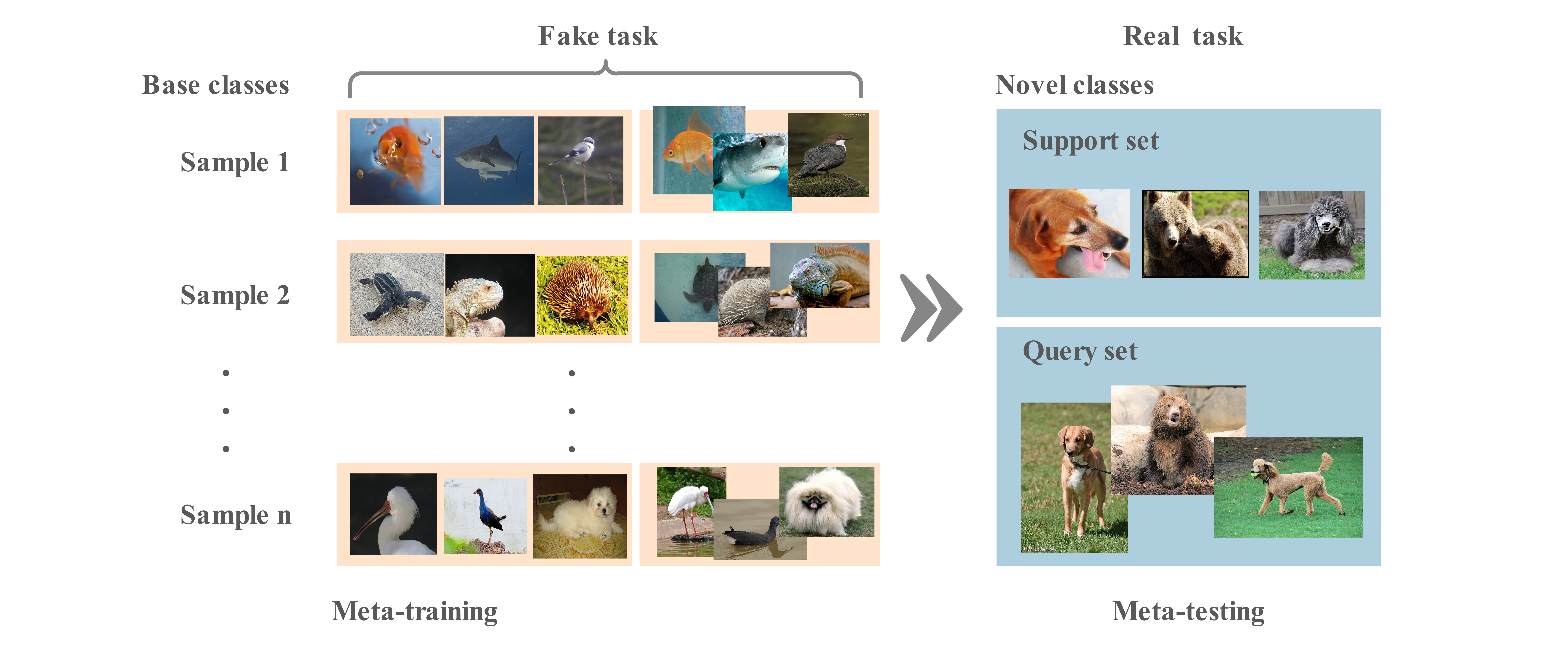}}
\caption{The universal method used in supervised few-shot learning, which consists of meta-training and meta-testing. In the meta-training, the training sample is actually a mimic few-shot task comprised of some labeled data chosen from base classes. And in the meta-testing the model will solve a real task with few labeled data and an unlabeled query set chosen from novel classes. We show a model trained for solving 3-way 1-shot tasks in this figure.}
\label{fig1}
\end{figure}

People have proposed varieties of few-shot methods, all of which rely on meta-training assisted with base classes. The universal approach is to use the base classes to construct fake few-shot tasks for training the network first, with the purpose of enabling the network an ability to accomplish real few-shot tasks through simulating the process of carrying out the fake tasks. This is called the meta-training stage with tasks as samples. Next, use the trained network to complete real few-shot tasks of novel classes, and calculate the classification accuracy on the query set in the tasks to evaluate the algorithm, which is usually called the meta-testing. The whole procedure is shown in Fig.\ref{fig1}.

Few-shot learning algorithms could be classified into three categories. The first \cite{vinyals2016matching,snell2017prototypical,sung2018learning,oreshkin2018tadam} is based on metric learning, which consists of three steps of feature extraction, distance measure, and prediction, relying on effective metric design and reducing the cross entropy loss in meta training to improve classification accuracy. The second are the teacher-student network based methods including \cite{finn2017model,rusu2018meta,li2019lgm,ravi2016optimization}. The teacher network guides the student network to solve the few-shot tasks in terms of parameter initialization, parameter update and other aspects. The algorithm enables the teacher network to obtain the ability to instruct the student network via meta-training. The third category such as \cite{liu2018learning} and \cite{kim2019edge} is based on the transduction, which propagates the label of the support data to the queries through a specific graph, thereby obtaining the predicted class of the query set. The algorithm optimizes the accuracy of propagation label through meta-training.

The meta-training determines the model’s performance in the few-shot learning algorithm. However it brings two obvious drawbacks. First, the meta-training requires a large number of labeled auxiliary examples (base classes). Those algorithms can not work without adequate labeled samples. Second, the meta-training phase uses tasks as training samples. Therefore, a task type decided by the values of N and k needs to be certain before meta-training to ensure that the number of images contained by each mimic few shot task (i.e. a meta-training sample) is consistent during training. The meta-trained network can only be used to solve few-shot tasks with the same type as the meta-training samples, and it performs worse in other types of tasks. However, in reality we need to solve various types of few-shot tasks, and it is unreasonable to meta-train the network from scratch in order to solve a few shot task with new type.

In order to solve these two problems, we propose a new paradigm of few shot learning based on contrastive self-supervised learning (CSSL-FSL). Specifically, our method abandons the meta-training phase, which takes the fake few-shot tasks as samples, and uses instead two new phases: the meta-training via self-supervised learning directly using a single image as a training sample, and the training of a classification network. In the first phase, a comparative self-supervised learning method is used to obtain a feature extractor with good generalization ability using unlabeled images. In the second phase, our method solve real few-shot tasks. The meta-trained feature extractor is used to extract features from all the images in the current task, and a feature aggregation is carried out, based on a specific graph defined by the current task so that the information of the query set can interact with that of the support set. We use the aggregated support set features to train a fully connected neural classification network. The classifier can predict classes of the query set after training.

Furthermore, we demonstrate that the self-distillation\cite{tian2020rethinking} and the manifold data augmentation are helpful for training the classification networks. Self-distillation is a special form of knowledge distillation used to promote classification accuracy. Manifold augmentation is used in our method to expand the training dataset by combining data at the semantic level.

Our key contributions can be summarized as follows:
\begin{itemize}
\item We propose CSSL-FSL, a new paradigm of unsupervised few-shot learning. By adopting the methodology of contrastive self-supervised learning, the two problems intrinsic in the existing meta-training paradigm are solved simultaneously. Thus our method does not require a large number of labeled samples for training. In addition, the meta-trained model can carry out different types of real few-shot tasks.
\item We leverage a graph with a trainable network parameter to aggregate features of samples in few-shot tasks to obtain more discriminating ones, which is similar to SGC \cite{wu2019simplifying}.
\item We propose to use the manifold augmentation and self-distillation technologies to alleviate the lack of labeled samples in the phase of training the classification network.
\item Adequate experiments demonstrate that our method reaches state-of-the-art accuracy on miniImageNet , a standardized benchmark in few-shot learning.
\end{itemize}

The paper is organized as follows. In \ref{Section 2}, we introduce the related works. Our methodology is described in \ref{Section 3}. In \ref{Section 4}, experimental results on the standard vision dataset are shown in comparison with the proposed works. Finally, a conclusion is drawn in \ref{Section 5}.

\section{Related Works}\label{Section 2}
In this section we aim to show the three types of supervised few-shot learning algorithms proposed in previous years. In addition, we introduce some unsupervised few-shot learning methods presented recently.

\subsection{Metric Based Methods}
The core of metric learning is to extract features from the support set and query set, then obtain the class prototypes using the support set, and predict classes of queries via the nearest neighbor algorithm and attention mechanism. Through meta-training, a metric based method obtains a feature extractor that facilitates completing the classification task based on distance measurement. Matching Networks \cite{vinyals2016matching} used LSTM to extract full context embeddings from images and applied attention mechanism to classify. Prototypical Networks \cite{snell2017prototypical} proposed to use Euclidean distance to better measure the similarity between features, and use prototypes of each class to classify queries. Relation Network \cite{sung2018learning} used a neural network to replace the traditional distance metric, and directly output the queries’ categories via an end-to-end network. DC-IMP \cite{lifchitz2019dense} introduced dense classification and leverage implanting to bring metric learning the task dependency.

\subsection{Teacher-Student Based Methods}
A Teacher-Student based method has two networks, called the teacher network and the student network. The student network is in charge of fulfilling the few-shot task, and the teacher network provide guidance on how to fulfil that task. A Teacher-Student based method ensures the teacher network possessing excellent guidance ability through meta-training, so that when facing real few-shot tasks, the teacher network can perform task-dependent guidance. In \cite{ravi2016optimization}, the teacher network guides on how the student network’s parameters update. When the student network updates, it does not obey the standard gradient descent, but uses the teacher network’s output as update values on the parameters. In MAML\cite{finn2017model}, the teacher network generates initial weights for the student network, which can help the student network converge quickly when facing new few-shot tasks. To further promote MAML, LEO \cite{rusu2018meta} proposed generating initial weights for student network from a lower-dimensional hidden space, which makes training more easily. The teacher network in LGM-Net \cite{li2019lgm} directly generates all the network parameters for its student network to deal with the few-shot task successfully. The student network does not need to finetune itself with support set, that is, the teacher network provides a one-step guidance.

\subsection{Transduction Based Methods}
The key of Transduction based methods is to integrate graphs into the algorithm. Through feature aggregation in a specific way, the features of the support set contain the information of the queries, which is equivalent to use dual information of support and query when updating the network. Through meta-training, transduction based methods can obtain excellent edge and vertex feature update modules. TPN \cite{liu2018learning} constructs a graph in the feature space, in which vertices are defined by image features and the adjacency matrix is obtained via calculating the vertices similarity. Then it initializes and updates a node-labeling matrix, and finally classifies the queries by updating the node-labeling matrix. EGNN \cite{kim2019edge} proposed a structure of graph similar to \cite{liu2018learning}, but used edge-labeling framework instead of node-labeling framework in classification which helps to exploit both the intra-cluster similarity and the inter-cluster dissimilarity.

\subsection{Unsupervised Methods}
The base classes in unsupervised methods has no labels. Some existing methods use unsupervised learning or data enhancement methods to leverage these unlabeled base classes to artificially construct fake support set and query set for meta training. They are able to combine with the few-shot learning methods as mentioned above (such as MAML \cite{finn2017model} and Prototypical Net \cite{snell2017prototypical}) to fulfil few-shot tasks. UFLST \cite{ji2019unsupervised} and CACTU \cite{hsu2018unsupervised} use clustering to make pseudo-labels for unlabeled examples, then use the pseudo-labeled data as ordinary labeled data to construct fake few-shot tasks to complete meta-training.  AAL \cite{antoniou2019assume} and UMTRA \cite{khodadadeh2019unsupervised} took each instance as one class and randomly sample multiple examples to construct a fake support set, then generate a corresponding query set according to the support set by data augmentation techniques. ULDA \cite{qin2020unsupervised} developed a new simple data augmentation method to enhance the difference between the support set distribution and query set distribution when constructing the fake few-shot tasks for meta-training.
\section{Methodology}\label{Section 3}
The notations and problem formulation of self-supervised few-shot learning are introduced in \ref{3.1}, and our paradigm is presented in \ref{3.2}. Finally, the self-knowledge distillation and manifold augmentation are described in \ref{3.3} and \ref{3.4} respectively. 
\subsection{Problem Formulation}\label{3.1}
Given two datasets, namely $D_{base}$ and $D_{task}$ with disjoint classes. $D_{base}$ consists of a large number of unlabeled examples from the base classes. $D_{task}$ has a small number of labeled examples called the support set $D_s$, along with some unlabeled ones called the query set $D_q$, all from the new classes. They stand for the total data in a few-shot learning task. The number of classes in the novel dataset $D_{task}$, the number of support samples and the number of query inputs for each of these classes are denoted $N$, $k$ and $q$ respectively. So there are totally $N\times(k+q)$ examples in a few-shot learning task. Our aim is to predict the classes of the query set of $D_{task}$. 
Different from the previous works like \cite{ravi2016optimization}, our $D_{base}$ does not have any labels. So we train the classification network with only a few labeled examples namely $D_s$ in a real sense.

\subsection{Proposed Paradigm of Solution}\label{3.2}
We first train a backbone deep neural network able to extract useful and compact features from inputs, which will be used as a generic feature extractor. In this so called meta-training phase, we train the network with $D_{base} = \left\{ x_{1}^{'},x_{2}^{'},......,x_{n}^{'} \right\}~$  where $x^{'} \in R^{w \times h \times 3}$ via CMC in \cite{tian2019contrastive}, a kind of contrastive self-supervised learning method,  which promises a transferable feature extractor. Thus we obtain the extractor $\left. f_{\varphi}:R^{w \times h \times 3}\rightarrow R^{2 \times e} \right.$ (consisting of two networks namely the $\left. f_{\varphi_{1}}:R^{w \times h}\rightarrow R^{e} \right.$  and $\left. f_{\varphi_{2}}:R^{w \times h \times 2}\rightarrow R^{e} \right.$, which will be described in detail later). 

We then use $f_\varphi$ to obtain the features of the total data in $D_{task}$ (both $D_s$ and $D_q$) namely $f_{\varphi}\left( D_{task} \right) = \left\{ f_{\varphi}\left( x \right) \middle| x \in D_{task} \right\}$. Then we step into the second phase namely the task-training phase. First we build a nearest neighbor graph using the cosine similarity according to \cite{hu2020exploiting} :
\begin{equation}
cos\left( {f_{\varphi}\left( x_{1} \right),f_{\varphi}\left( x_{2} \right)} \right) = \frac{f_{\varphi}\left( x_{1} \right)^{T}f_{\varphi}\left( x_{2} \right)}{\left\| {f_{\varphi}\left\| x_{1} \right\|} \right\|_{2}\left\| {f_{\varphi}\left\| x_{2} \right\|} \right\|_{2}}
\end{equation}
The base graph denoted $G_{task}\left( V,E \right)$ uses $f_\varphi(D_{task})$ to construct vertices. In details, its vertices matrix $V \in R^{\lbrack N \times (k + q)\rbrack \times 2e}$ is the stacked representations of support set and query set i.e. each vertex represents an image's feature. We make the values of graph edges represent the similarity between vertices-that is, similar vertices have larger adjacency values. To get the adjacency matrix $E \in R^{\lbrack N \times (k + q)\rbrack \times \lbrack N \times (k + q)\rbrack}$, we first define a similarity matrix $S$ with the same dimension computed as follows:
\begin{equation}
S_{i,j} = \begin{cases}
\left. {\mathit{\cos}(}V_{i,:},V_{j,:} \right) & {i \neq j} \\
0 & {i = j} \\
\end{cases}
\end{equation}
where $V_{i,:}$ denotes the $i$ -th row in $V$. Then we just save the $m$ largest values on each row and on the corresponding column in S to obtain a more sparse matrix helpful to reduce the interference. Finally, we normalize the resulting matrix to get the adjacency matrix:
\begin{equation}
E = D^{- \frac{1}{2}}SD^{- \frac{1}{2}}
\end{equation}
where $D$ is the degree diagonal matrix computed by $D_{i,i}=\sum_{j} S_{i,j}$. We can consider $E$ as the Laplacian matrix in GCN\cite{kipf2016semi} used to aggregate information among vertices.

Then we aggregate features for each vertex via the graph structure to get $V^{new}$ and train a fully connected network $Cls_{\theta}{{:R}^{2e}\rightarrow}R^{N}$ using vertices defined by $D_s$. During the training we take two sub-stages to achieve a better performance with quite few support examples. In the two sub-stages, we use manifold augmentation from which we can obtain augmented data $mixed{\_ V}_{i,:}^{new}$ and self-knowledge distillation to alleviate overfitting and to achieve a better performance respectively. 

\begin{figure*}[htbp]
\centering
\includegraphics[width=\linewidth,scale=1.00]{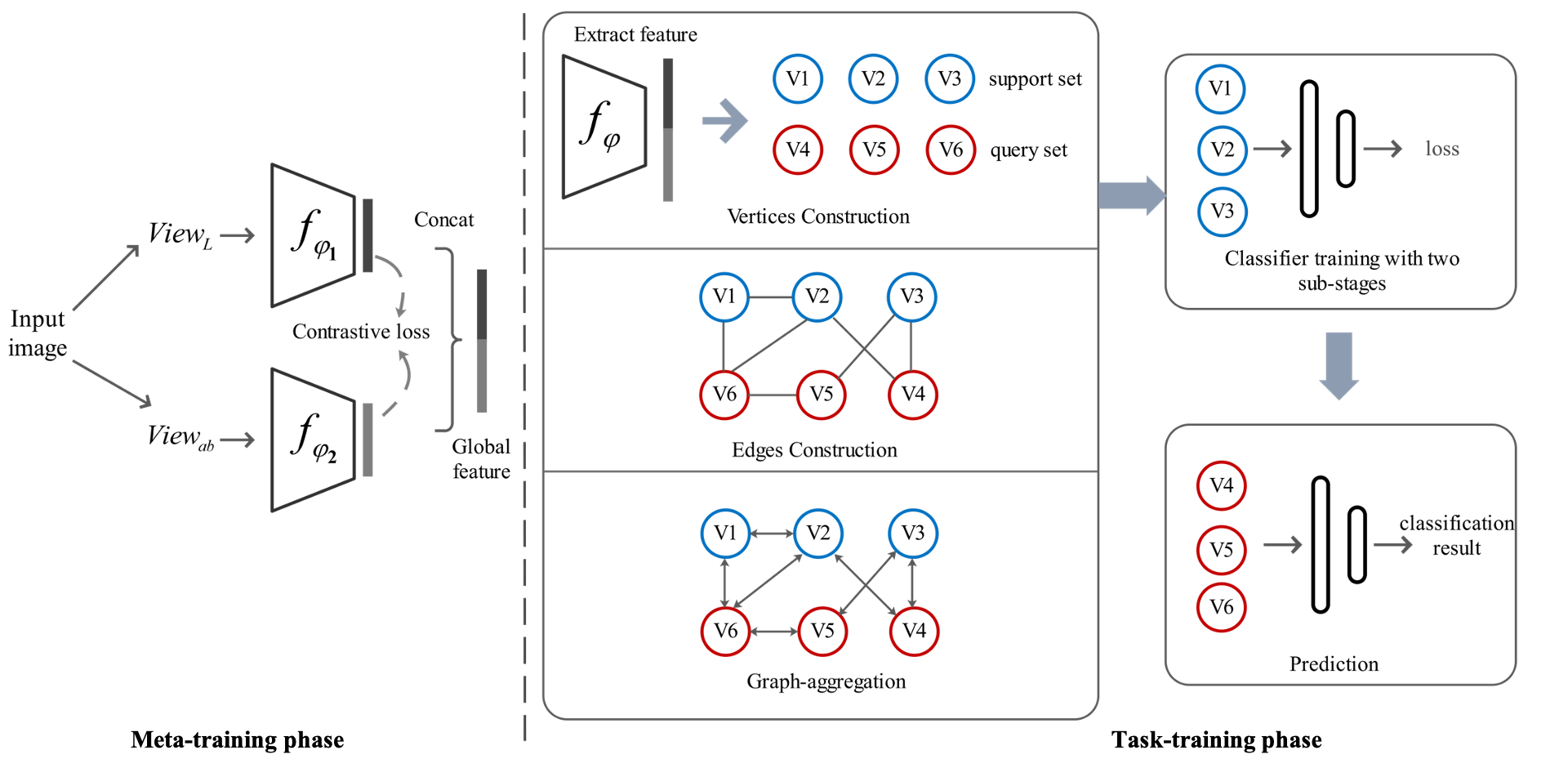}
\caption{The overall architecture of the proposed paradigm. The left shows the meta-training phase using CMC resulting in a global feature extractor. The right is the task-training phase comprised of graph-aggregation, classifier training with the support set and predition on the query set.}
\label{fig2}
\end{figure*}

\begin{algorithm}[ht]
\caption{The process of meta-training phase.}
\label{algorithm1}
\KwIn{$D_{base} = \left\{ x_{1}^{'},x_{2}^{'},......,x_{n}^{'} \right\}$  without labels; }
\KwOut{A meta-trained feature extractor $f_\varphi$ consists in $f_{\varphi_1}$ and $f_{\varphi_2}$}
initialize $f_\varphi$ \;
\If{training is not completed}
{
Choose a minibatch $D_{batch}$ from $D_{base}$ randomly\;
Feed $D_{batch}$ into $f_\varphi$ to obtain $f_\varphi(D_{batch})$\;
Compute contrastive loss $L_{SSL}$ using $f_\varphi(D_{batch})$ according to CMC \cite{tian2019contrastive}\;
Update $\varphi$ with $L_{SSL}$\;
}
\textbf{Return} $f_{\varphi}$;

\end{algorithm}

% \renewcommand{\algorithmicrequire}{\textbf{Input:}} 
% \renewcommand{\algorithmicensure}{\textbf{Output:}} 
% \begin{algorithm}[htb] 
% \caption{The process of pre-training phase.} 
% \label{algorithm1} 
% \begin{algorithmic}[1] 
% \Require 
% $D_{base} = \left\{ x_{1}^{'},x_{2}^{'},......,x_{n}^{'} \right\}$  without labels; 
% \Ensure 
% A pretrained feature extractor $f_\varphi$ consists in $f_{\varphi_1}$ and $f_{\varphi_2}$; 
% \REPEAT
% \STATE Choose a minibatch $D_{batch}$ from $D_{base}$ randomly;
% \STATE Feed $D_{batch}$ into $f_\varphi$ to obtain $f_\varphi(D_{batch})$;
% \STATE Compute contrastive loss $L_{SSL}$ using $f_\varphi(D_{batch})$ according to CMC \cite{};
% \STATE Update $\varphi$ with $L_{SSL}$; 
% \UNTIL{Training completed}
% \RETURN $ f_{\varphi}$ 
% \end{algorithmic}
% \end{algorithm}

\begin{algorithm}[ht]
\caption{The process of task-training phase with two sub-stages.}
\label{algorithm2}
\KwIn{A $N$ -way $k$ -shot task with the dataset $D_{task}=\left\{ D_{s},D_{q} \right\}$; The meta-trained feature extractor $f_\varphi$ consists in $f_{\varphi_1}$ and $f_{\varphi_2}$}
\KwOut{Parameters for the classifier $Cls_\theta$}
Obtain features of all inputs including labeled and unlabeled ones, $f_{\varphi}\left( D_{task} \right)$\;
Build the graph $G_{task}(V,E)$ based on $f_\varphi(D_{task})$\;
Aggregate vertices features $V$ of the graph to get $V^{new}$\;
randomly initialize $\theta$; \qquad\qquad\qquad\qquad\qquad\qquad\textbf{The first sub-stage}\;
Use manifold augmentation to extend labeled data in semantic level, and get the augmented feature set $V_{aug}^{new} = \left\{ mixed{{\_ V}^{\mathit{new}},}V^{new} \right\}$\;
Train $Cls_\theta$ using $V_{aug}^{new}$ and cross entropy loss to obtain $Cls_{\theta_0}$;\qquad\qquad\qquad\qquad\qquad\qquad\qquad\qquad\textbf{The second sub-stage}\;
Use the predictions on $D_s$ from $Cls_{\theta_0}$ and labels of $D_s$ to compute distillation loss\;
Update $\theta$ from scratch and finally obtain $\theta_1$\;
\textbf{Return} $ Cls_{\theta_1}$;

\end{algorithm}

% \renewcommand{\algorithmicrequire}{\textbf{Input:}} % Use Input in the format of Algorithm
% \renewcommand{\algorithmicensure}{\textbf{Output:}} % Use Output in the format of Algorithm
% \begin{algorithm}[htb] 
% \caption{The process of training phase with two sub-stages.} 
% \label{algorithm2} 
% \begin{algorithmic}[1] 
% \Require 
% A $N$ -way $k$ -shot task with the dataset $D_{task}=\left\{ D_{s},D_{q} \right\}$; 
% The pretrained feature extractor $f_\varphi$ consists in $f_{\varphi_1}$ and $f_{\varphi_2}$; 
% \Ensure 
% Parameters for the classifier $Cls_\theta$; 
% \STATE Obtain features of all inputs including labeled and unlabeled ones, $f_{\varphi}\left( D_{task} \right)$; 
% \STATE Build the graph $G_{task}(V,E)$ based on $f_\varphi(D_{task})$; 
% \STATE Aggregate vertices features $V$ of the graph to get $V^{new}$; 
% \STATE randomly initialize $\theta$; \\
% \textbf{The first sub-stage:}
% \qquad\STATE Use manifold augmentation to extend labeled data in semantic level, and get the augmented feature set $V_{aug}^{new} = \left\{ mixed{{\_ V}^{\mathit{new}},}V^{new} \right\}$
% \qquad\STATE Train $Cls_\theta$ using $V_{aug}^{new}$ and cross entropy loss to obtain $Cls_{\theta_0}$;\\
% \textbf{The second sub-stage (self-distillation):}
% \qquad\STATE Use the predictions on $D_s$ from $Cls_{\theta_0}$ and labels of $D_s$ to compute distillation loss;
% \qquad\STATE Update $\theta$ from scratch and finally obtain $\theta_1$;
% \RETURN $ Cls_{\theta_1}$
% \end{algorithmic}
% \end{algorithm}

Our paradigm is illustrated in Fig.\ref{fig2}. In general CSSL-FSL has two phases: (1) Meta-training phase: training a generic feature extractor via contrastive self-supervised learning. (2) Task-training phase: adapting a classification network using the support set data after the feature aggregation through graph. Once the latter is finished, the performance of this model is evaluated on the vertices constructed from $D_q$. The process of the meta-training phase is provided in Algorithm \ref{algorithm1} and the process of task-training phase is formalized in Algorithm \ref{algorithm2}.

The details of the two phases are provided in the following, ﬁrst the meta-training phase then the task-training phase.

\textbf{Meta-training phase:} We follow the methodology called CMC, an effective contrastive self-supervised learning method, proposed in \cite{tian2019contrastive}. More specifically, we consider an input image in Lab color space, spliting it into $L$ view (luminance) called $View_L$ and the $ab$ view (chrominance) called $View_{ab}$. We aim to obtain a network able to extract compact and distinct features from the inputs. Through contrastive learning we learn a feature embedding, which can map views of similar images to nearby points while map views of different images to far apart points. The feature embedding have two parts, the $L$ view part termed $f_{\varphi_1}$ and the $ab$ view part $f_{\varphi_2}$. So we have $\varphi=(\varphi_1,\varphi_2)$. The total feature of the input is the concatenation of the outputs from these two parts, namely $f_\varphi(x)=concat[f_{\varphi_1}(x),f_{\varphi_2}(x)]$.

As in \cite{tian2019contrastive} we use the contrast loss as a loss for the self-supervised learning to train the feature embedding:
\begin{equation}
L_{SSL} = L_{contrast}^{View_{L},View_{ab}} + L_{contrast}^{View_{ab},View_{L}}
\end{equation}
in which $L_{contrast}^{View_L,View_{ab}}$ is the contrast loss computed by treating view $View_L$ as anchor and enumerates over $View_{ab}$ while $L_{contrast}^{View_{ab},View_L}$ anchors at $View_{ab}$.

\textbf{Task-training phase:} We ﬁx the meta-trained parameters $\varphi = \left\{ \varphi_{1}\text{,~}\varphi_{2} \right\}$ in the backbone and train a task-dependent classiﬁer $Cls_\theta$ on the transferred representations of the few-shot task’s dataset namely $D_{task}$. Before training a linear classiﬁer with $D_s$ having few labeled examples, a method similar to simpliﬁed graph convolution \cite{wu2019simplifying}, namely the graph aggregation is used. We construct the graph $G_{task}(V,E)$ for the current few-shot task through the steps introduced before. The vertices in $G_{task}$ contain the total representations of $D_{task}$ and the adjacency matrix stands for the coefficient of attention used in features aggregation.

We then propagate feature (\cite{wu2019simplifying}) to obtain new features for each vertex:
\begin{equation}
V^{new} = \left( \alpha I + E)^{\gamma}V \right.
\end{equation}
where $I$ is the identity matrix and $\gamma$ is a hyperparameter which plays an important role in getting better representation, denoting the number of times to aggregate feature. At the same time, $\alpha$ is also a key value to balance between the neighbors representations and the self-ones. So we make it a trainable network parameter instead of a fixed value as in SGC.

After aggregation, we use the labeled part of the vertices to train the task-dependent classifier $Cls_\theta$, a simple fully connected network. We may choose to extend the support set by manifold augmentation and train with the cross entropy loss, and we could use the self-distillation to further improve the performance, which are introduced in details in \ref{3.3} and \ref{3.4} respectively.

\subsection{Manifold Augmentation}\label{3.3}
Since the number of labeled samples in a few-shot task is too small, in the first sub-stage of the classifier training we use data augmentation to expand the training set. Manifold Mixup\cite{verma2018manifold} is a kind of effective data augmentation method, which leverages semantic interpolations as additional training signal along with the corresponding linearly combined labels. It was proved that the combination of hidden representations of training examples works better than the original image mixup.

In our method, we use Manifold Mixup based on the new vertices matrix as follows:
\begin{equation}
mixed{{\_ V}_{i,:}^{\mathit{new}} =}\lambda V_{i,:}^{new} + \left( 1 - \lambda \right)V_{j,:}^{new}~~1 \leq i \leq N \times k
\end{equation}
where $V_{i,:}^{new}$ is the $i$ -th row in the new vertices matrix as the base feature, $V_{j,:}^{new}$ plays as noise ($j$ is randomly selected from $\left[1,N\times k\right]$), and $mixed{\_ V}_{i,:}^{new}$ denotes the augmented embedding based on $V_{i,:}^{new}$. We make $\lambda$ close to one to ensure that our base embedding won’t get much change because we still use the original label of the base feature for the combined feature. We also tried to use the linear combined labels but it leads to worse performance. 
Then we leverage both the original and the augmented vertices to update both the parameters $\theta$ in $Cls$ and $\alpha$ used in graph aggregation with cross entropy loss. We regard $\alpha$ as a part of $\theta$ for convenience in the following.

\subsection{Self-distillation}\label{3.4}
\begin{figure}[htbp]
\centerline{\includegraphics[width=8cm, height=2.35cm]{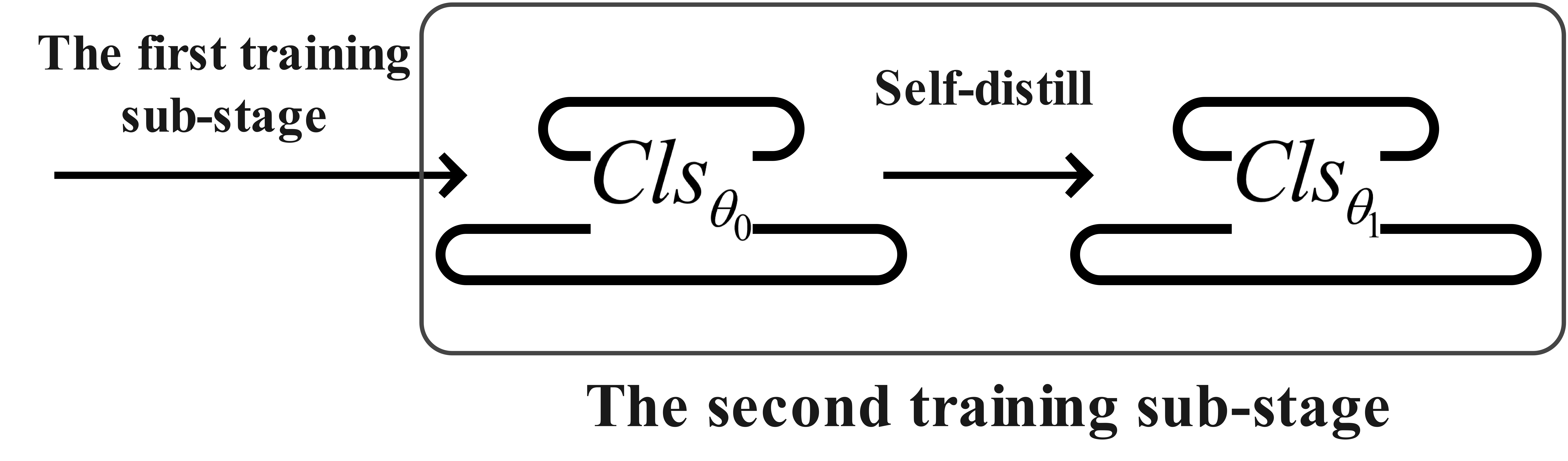}}
\caption{In the second training sub-stage, $Cls_{\theta_1}$ is learned with knowledge distilled from $Cls_{\theta_0}$ which has been trained in the first sub-stage.}
\label{fig3}
\end{figure}

Knowledge distillation \cite{yim2017gift} is usually used to get a compact network called student network, which leverages a complex but well-performed teacher network to get soft-targets as auxiliary label during training. Distillation can be seen as a method of knowledge transfer. In the second stage of classifier training, the self-distillation is used, in other words, the student network and teacher network have the same architecture. We term the classifier produced by first training sub-stage $Cls_{\theta_0}$ with parameters $\theta_0$, which works as the teacher network in the second training sub-stage. And $Cls_{\theta_1}$ is the student network obtained from self-distillation with parameters $\theta_1$, which is shown in Fig.\ref{fig3}. In this sub-stage we only use the original labeled vertices (no mixed vertices) as training data.

We use convex combination of the cross-entropy loss between the predictions and the one-hot labels and the Kullback Leibler divergence (KL) between predictions and soft targets predicted by $Cls_{\theta_0}$ as loss function:

\begin{equation}
\begin{split}
L_{distill} = \frac{1}{N \times k}{\sum_{i = 1}^{N \times k}{\beta L^{ce}\left( cls\left( V_{i,:}^{new}{;\theta} \right),gt_{i} \right)}} \\
+ \left( 1 - \beta \right)KL\left( cls\left( V_{i,:}^{new}{;\theta} \right),cls\left( V_{i,:}^{new}{{;\theta}_{0})} \right) \right.
\end{split}
\end{equation}
where $gt_i$ is the groundtruth of the $i$–th vertex. $Cls_{\theta_1}$ is the final classifier to predict the class of query set in current few-shot task and we use unlabeled vertices get from aggregated vertices matrix $V^{new}$ as classifier’s inputs to obtain predictions on the query set.

\section{Experiments}\label{Section 4}

We conduct experiments on the widely used few-shot image classification benchmark: miniImageNet \cite{vinyals2016matching}, which is a derivative of ImageNet. 
\subsection{Models and implementation details}
\textbf{Architecture.} In the meta-training phase we use ResNet50-v1\cite{he2016deep} as the structure of feature extractors $f_{\varphi_1}$ and $f_{\varphi_2}$. This backbone has 50 convolutional layers grouped into 16 blocks. ‘-v1’ means the width of each of the convolutional layers is half of the original ResNet50. We set the input size as $224\times224$ and flatten the outputs after the average-pooling layer as inputs to the graph aggregation, so that $e=1024$ in $V \in R^{\lbrack N \times (k + q)\rbrack \times 2e}$.

In consideration of the extreme few labeled examples we take only one fully connected layer and a following softmax layer as the structure of the classifier $Cls$ to avoid overfitting.

\textbf{Optimization and hyper-parameters setup.} For the meta-training phase, we train the backbone in a total of 240 epochs from scratch using the SGD optimizer \cite{bottou2010large} and the contrast loss. For the task-training phase, we expand labeled vertices 120 times by manifold augmentation and in the first and the second training sub-stage we train the classifier in 11 epochs, 1000 epochs respectively, using the Adam optimizer \cite{kingma2014adam} and the loss function shown in the previous section. In the second training sub-stage we set $\lambda$ as 0.95. 

\begin{table}[htbp]
\caption{Performance of CSSL-FSL in comparison to the previous works on miniImageNet on 5-way 1-shot and 5-way 5-shot tasks. Average accuracies are reported with 95\% conﬁdence intervals.}
\begin{center}
\begin{tabular}{p{1.2cm}p{3.4cm}p{1.4cm}<{\centering}p{1.1cm}<{\centering}}
\hline
\multicolumn{2}{c}{}&\multicolumn{2}{p{2.5cm}<{\centering}}{\textbf{5-way Accuracy}} \\
\hline
\multicolumn{2}{c}{\textbf{miniImageNet}}& \textbf{1-shot} &\textbf{5-shot} \\
\hline
&\textbf{CACTUs-MAML \cite{hsu2018unsupervised}}	&39.90$\pm$0.74\%	&53.97$\pm$0.70\% \\
&\textbf{CACTUs-ProtoNets \cite{hsu2018unsupervised}}&	39.18$\pm$0.71\%	&53.36$\pm$0.70\%\\
&\textbf{UFLST \cite{ji2019unsupervised}} &33.77$\pm$0.70\%	&45.03$\pm$0.73\%\\
% &\textbf{UMTRA \cite{khodadadeh2019unsupervised}} &39.93$\pm$-\% &50.97$\pm$-\% \\
&\textbf{UMTRA \cite{khodadadeh2019unsupervised}} &39.93$\pm-$\%	&50.73${\pm-}$\% \\
\multirow{2}{*}{\textbf{unsupervised}} & \textbf{AAL-ProtoNets \cite{antoniou2019assume}}	&37.67$\pm$0.39\%	&40.29$\pm$0.68\%\\
&\textbf{AAL-MAML++ \cite{antoniou2019assume}}	&34.57$\pm$0.74\%	&49.18$\pm$0.47\%\\
&\textbf{ULDA-ProtoNets \cite{qin2020unsupervised}}	&40.63$\pm$0.61\%	&55.41$\pm$0.57\%\\
&\textbf{ULDA-MetaOptNet \cite{qin2020unsupervised}}	&40.71$\pm$0.62\%	&54.49$\pm$0.58\%\\
&\textbf{CSSL-FSL\_Mini64(ours)}	&\textbf{48.53$\pm$1.26\%}	&63.13$\pm$0.87\%\\
&\textbf{CSSL-FSL\_Image168(ours)}	&\textbf{54.17$\pm$1.31\%}	&\textbf{68.91$\pm$0.90\%}\\
\hline
\multirow{2}{*}{\textbf{supervised}} &\textbf{MAML}	&46.60$\pm$0.74\%	&60.00$\pm$0.71\%\\
&\textbf{ProtoNets}	&47.01$\pm$0.72\%	&67.90$\pm$0.76\%\\
\hline
\multicolumn{4}{p{8.7cm}}{‘\_Mini64’ means pretrain on the base classes in miniImageNet, ‘\_Image168’ means pretrain on the larger dataset we chose.}
\end{tabular}
\label{tab1}
\end{center}
\end{table}

\subsection{Results on miniImageNet}
The miniImageNet dataset consists of 100 classes randomly sampled from the ImageNet and each class contains 600 images of size $84\times84$. It is usually divided into three parts \cite{ravi2016optimization}: training set with 64 base classes, validation set with 16 classes, and testing set with 20 novel classes. In the meta-training phase we use 64 base classes without labels as a small training dataset and 168 classes randomly chosen from ImageNet by ourselves as a bigger one, also having no labels. In the task-training phase we sample novel classes to design few-shot tasks as inputs. We ensure that the novel classes have never been seen in the meta-training phase.

We evaluate our method on 600 randomly sampled tasks and report their mean accuracy in TABLE \ref{tab1}. We compare our method in both 5-way 1-shot and 5-way 5-shot setting with some classical supervised few-shot learning methods and novel unsupervised methods proposed recently. It can be found that our method is much better than previous unsupervised few-shot learning methods( \cite{ji2019unsupervised} etc.), improving them by more than 10\%. Even compared with supervised methods(\cite{finn2017model} and \cite{snell2017prototypical}), our method still has improvement by 1-8\% when pretrain with a larger dataset both on 5-way 1-shot and 5-way 5-shot tasks.

We notice that using larger training set in meta-training phase leads to an obvious improvement by 5-6\%. This is in accordance with the property of contrastive self-supervised learning. When the backbone has seen more images, it can extract features better. So we believe that our method can achieve a better performance by further extending the meta-training dataset. And in the following we show the results of experiments with larger meta-training dataset.
\subsection{Results on multi-type tasks}
To prove that our method can easily generalize to different types of few-shot tasks after only pretrainig once, we show the results compared with EGNN \cite{kim2019edge}, a method needing meta-training based on fixed-type fake tasks, in TABLE \ref{tab2}. For the meta-training in EGNN we use 5-way 5-shot fake tasks as training set, and then we evaluate it on multi-type tasks. For our method we directly use the meta-trained backbone to solve few-shot tasks with different $k$. This setting ensures that both methods have just one meta-training process.

When the number of support examples in a few-shot task increases, the model should perform better because the more labeled data usually leads to better generalization. So it is not reasonable that when $k$ increases to 20, EGNN gets a worse performance. However the accuracy of our method keeps rising with the growth of $k$. Our method outperforms EGNN by 3\% when $k=20$ and improves further by 7\% when $k=30$. The results show that, compared to the previous works based on meta-training, our method is not limited by the type of tasks, and it only needs one meta-training to obtain a model with outstanding performance of generalization.
\begin{table}[htbp]
\caption{Performance of CSSL-FSL in comparison to EGNN on miniImageNet on different types of tasks. For each type of task, the best-performing method is in bold.}
\begin{center}
\begin{tabular}{l p{0.7cm}<{\centering} p{0.7cm}<{\centering} p{0.8cm}<{\centering} p{0.8cm}<{\centering} p{0.8cm}<{\centering}}
\hline
\textbf{5-way Accuracy}&\textbf{1shot}&\textbf{5shot}&\textbf{10shot} &\textbf{20shot} &\textbf{30shot} \\
\hline
\textbf{EGNN} &	44.74\%	&\textbf{76.30\%}	&\textbf{77.40\%}	&75.49\%	&72.83\% \\
\textbf{CSSL-FSL\_Image168}	&\textbf{54.17\%}	&68.91\%	&74.82\%	&\textbf{78.47\%}	&\textbf{80.83\%} \\
\hline
\end{tabular}
\label{tab2}
\end{center}
\end{table}

\subsection{Ablation experiments}
In this section, we conduct ablation experiments to analyze how self-distillation and manifold augmentation affects the few-shot image classification performance. 
TABLE \ref{tab3} shows the results of the ablation studies on miniImageNet in 5-way 1-shot and 5-way 5-shot setting. We compare three kinds of ablation models in the following: (1) \textbf{w/o distill}: Don’t use self-distillation, meaning that classifier is just trained with the first sub-stage. (2) \textbf{w/o aug}: This is the model without manifold augmentation in the fist sub-stage but it still have self-distillation in the second sub-stage. (3) \textbf{w/o both}: This ablation model just has the first training sub-stage without manifold augmentation. Overall, the original model performs best. Self-distillation improves accuracy by 0.4-0.9\%. Manifold augmentation can provide 0.8-2\% extra gain. And without both of distillation and augmentation, the result in 1-shot case is greatly affected, decreasing by 3\%.

\begin{table}[htbp]
\caption{Results of ablation studies on miniImageNet. The meta-training dataset consists of 168 classes from ImageNet for all the four models.}
\begin{center}
\begin{tabular}{l c c}
\hline
\textbf{5-way Accuracy}&\textbf{1-shot}&\textbf{5-shot} \\
\hline
\textbf{CSSL-FSL -w/o distill} &53.28$\pm$1.01\% &68.25$\pm$0.91\% \\
\textbf{CSSL-FSL-w/o aug} &51.92$\pm$1.04\%	&67.94$\pm$0.85\% \\
\textbf{CSSL-FSL-w/o both}	&50.99$\pm$1.05\%	&67.93$\pm$0.85\% \\
\textbf{CSSL-FSL\_Image168}	&\textbf{54.17$\pm$1.31\%}	&\textbf{68.91$\pm$0.90\%} \\
\hline
\end{tabular}
\label{tab3}
\end{center}
\end{table}

\section{Conclusion}\label{Section 5}
A novel paradigm of unsupervised few-shot learning is proposed in this paper, which consists of two phases: the contrastive self-supervised learning to obtain a transferable feature extractor, and the graph-aggregation followed with classifier training.

Experiments show a state-of-the-art performance on a standard vision dataset miniImageNet. It proves that, without a large number of labeled data an outstanding backbone can still be obtained to extract transferable features. And with just one model meta-training, different types of few-shot tasks can be achieved even better. This paradigm can be used in different areas other than the image classification. In the following study we will explore efficient ways in defining the graph and aggregating the features.

% \subsection{Figures and Tables}
% \paragraph{Positioning Figures and Tables} Place figures and tables at the top and 
% bottom of columns. Avoid placing them in the middle of columns. Large 
% figures and tables may span across both columns. Figure captions should be 
% below the figures; table heads should appear above the tables. Insert 
% figures and tables after they are cited in the text. Use the abbreviation 
% ``Fig.~\ref{fig}'', even at the beginning of a sentence.

% \begin{table}[htbp]
% \caption{Table Type Styles}
% \begin{center}
% \begin{tabular}{|c|c|c|c|}
% \hline
% \textbf{Table}&\multicolumn{3}{|c|}{\textbf{Table Column Head}} \\
% \cline{2-4} 
% \textbf{Head} & \textbf{\textit{Table column subhead}}& \textbf{\textit{Subhead}}& \textbf{\textit{Subhead}} \\
% \hline
% copy& More table copy$^{\mathrm{a}}$& &  \\
% \hline
% \multicolumn{4}{l}{$^{\mathrm{a}}$Sample of a Table footnote.}
% \end{tabular}
% \label{tab}
% \end{center}
% \end{table}

% \section*{Acknowledgment}

% The preferred spelling of the word ``acknowledgment'' in America is without 
% an ``e'' after the ``g''. Avoid the stilted expression ``one of us (R. B. 
% G.) thanks $\ldots$''. Instead, try ``R. B. G. thanks$\ldots$''. Put sponsor 
% acknowledgments in the unnumbered footnote on the first page.

% \section*{References}
\bibliographystyle{IEEEtran}
\bibliography{Few-Shot_Image_Classification_via_Contrastive_Self-Supervised_Learning}
\vspace{12pt}
\end{document}